\documentclass[letterpaper,10 pt,conference]{ieeeconf}  %
\usepackage{threeparttable}
\usepackage{graphicx}
\usepackage{subfigure}
\usepackage{xcolor}
\usepackage{amssymb}
\usepackage{footnote}
\usepackage{dsfont}

\IEEEoverridecommandlockouts                              %

\usepackage{amsmath} %
\usepackage{amssymb}  %
\usepackage{caption}
\usepackage{bbm}

\title{\LARGE \bf
DORec: Decomposed Object Reconstruction and Segmentation Utilizing 2D Self-Supervised Features
}

\author{Jun Wu, Sicheng Li, Sihui Ji, Yifei Yang, Yue Wang, Rong Xiong, and Yiyi Liao$^{\dagger}$
\thanks{$^{\dagger}$Corresponding author {\tt\small yiyi.liao@zju.edu.cn}}%
\thanks{Jun Wu, Sicheng Li, Sihui Ji, Yifei Yang, Yue Wang, Rong Xiong and Yiyi Liao are with Zhejiang University, Hangzhou, China. }
}

\newcommand{\bc}{\mathbf{c}}\newcommand{\bC}{\mathbf{C}}
\newcommand{\bd}{\mathbf{d}}

\newcommand{\bI}{\mathbf{I}}

\newcommand{\bn}{\mathbf{n}}
\newcommand{\bo}{\mathbf{o}}
\newcommand{\bp}{\mathbf{p}}

\newcommand{\br}{\mathbf{r}}
\newcommand{\bs}{\mathbf{s}}

\newcommand{\bx}{\mathbf{x}}

\newcommand{\bxi}{\boldsymbol{\xi}}

\newcommand{\cL}{\mathcal{L}}

\newcommand{\cS}{\mathcal{S}}

\makeatletter
\DeclareRobustCommand\onedot{\futurelet\@let@token\@onedot}
\def\@onedot{\ifx\@let@token.\else.\null\fi\xspace}

\makeatother

\definecolor{darkgreen}{rgb}{0,0.7,0}

\begin{document}

\input{teaser}
\maketitle

\begin{abstract}
Recovering 3D geometry and textures of individual objects is crucial for many robotics applications, such as manipulation, pose estimation, and autonomous driving. 
However, decomposing a target object from a complex background is challenging. 
Most existing approaches rely on costly manual labels to acquire object instance perception. 
Recent advancements in 2D self-supervised learning offer new prospects for 
identifying objects of interest, yet leveraging such noisy 2D features for clean decomposition remains difficult. 
In this paper, we propose a Decomposed Object Reconstruction (DORec) network based on neural implicit representations. 
Our key idea is to use 2D self-supervised features to create two levels of masks for supervision: a binary mask for foreground regions and a K-cluster mask for semantically similar regions. 
These complementary masks result in robust decomposition. 
Experimental results on different datasets show DORec's superiority in segmenting and reconstructing diverse foreground objects from varied backgrounds enabling downstream tasks such as pose estimation.
\end{abstract}
\section{INTRODUCTION}

Reconstructing a 3D object from multi-view images aims to obtain accurate and compact geometry and photo-realistic textures. 
Decomposing the 3D object from the background is crucial for 3D scene understanding and can be applied to various robotic applications, such as manipulation, pose estimation, and autonomous driving (Fig.~\ref{fig:teasesr}). 
For example, a decomposed 3D reconstruction can replace known CAD models in object pose estimation. 
Recently, neural implicit representations~\cite{park2019deepsdf,mescheder2019occupancy} greatly enhance reconstruction quality, but decomposing a target object from the background without costly manual annotation remains challenging.

To address this,
\cite{zhi2021place,wu2022object} use 2D masks obtained from pre-trained segmentation models to avoid expensive manual annotation. 
However, these coarse-grained masks are prone to errors, affecting the precision of decomposed modeling (see Fig.~\ref{fig:teasesr}, SemanticNeRF). 
Other approaches exploit fine-grained features from 2D self-supervised models for decomposed reconstruction, distilling these features into the 3D space for post-processing decomposition~\cite{yu2021unsupervised}, e.g., identifying a specific 3D region based on the feature vector of a user query. 
While promising, this makes them susceptible to noise and false classifications (Fig.~\ref{fig:teasesr}, DFF).

In this paper, we propose utilizing self-supervised 2D features of varying granularity for decomposed object reconstruction. 
Our key idea is to transfer these features into two types of masks: a coarse-grained mask for rough decomposition and a median-grained mask for noise resistance. 
Self-supervised features with median granularity, such as a K-cluster segmentation mask, are less prone to erroneous classification than binary masks (Fig.~\ref{fig:teasesr}). 
Median-grained clustering can be easily used during training to encourage decomposed reconstruction, avoiding post-processing region extraction. 
However, using only median-grained features might lead to alternative solutions due to the lack of clear foreground-background guidance. 
Therefore, we also use coarse-grained masks to provide distinct boundary indications, ensuring reliable convergence during early training stages.
For reconstruction, we employ a compositional network, with the foreground model as a neural surface representation for high-quality geometries and the background model as a neural radiance field for unbounded scenes.
The network is supervised by RGB reconstruction loss and segmentation losses using both coarse-grained and median-grained masks. 
Our proposed training strategy effectively leverages both masks for decomposition. 
After training, our model can render accurate foreground object masks and extract the foreground object's surface.

\begin{figure*}[htb]
    \centering
        \includegraphics[width=0.9\linewidth]{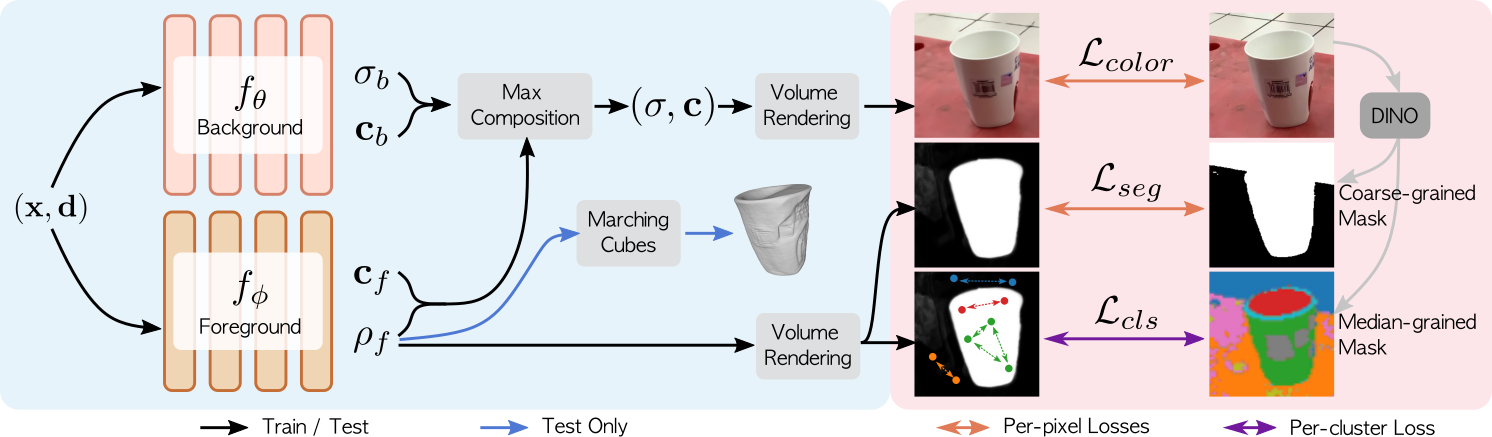}
   \caption{\textbf{Method overview}. The left part in blue shows our decomposed network consisting of a background model $f_\theta$ and a foreground model $f_\phi$. The background predictions $(\bc_b,\sigma_b)$ and foreground predictions $(\bc_f, \rho_f)$ are blended together via point-wise max composition. The right part in pink illustrates how we leverage self-supervised 2D features of varying granularity to enable decomposed reconstruction. 
   }
\label{fig:overview}
\vspace{-0.7cm} 
\end{figure*}

In summary, the major contributions of this work are:
i) To address the decomposed object reconstruction problem, we propose a compositional neural implicit network that uses 2D self-supervised features for decomposition, sparing pixel-wise annotations.
ii) We propose to utilize self-supervised features of varying granularity, with coarse-grained features for early guidance and median-grained features for noise resilience.
iii) Our method obtains high reconstruction quality and neat segmentation results, as proved by experiments on various object-centered and cluttered real datasets.
\section{Related Works}

\subsection{Neural Implicit Representations}
Traditional object reconstruction methods often use discrete representations, including voxel-based~\cite{choy20163d}, point-based~\cite{qi2017pointnet} and mesh-based~\cite{groueix2018papier}. Neural implicit representations~\cite{mescheder2019occupancy,park2019deepsdf} consider the 3D space as a continuous function by mapping each 3D point to a signed distance field (SDF) or an occupancy value, enabling high reconstruction accuracy and thus gain popularity~\cite{yariv2020multiview,liu2020neural,niemeyer2020differentiable}. 
However, these methods require 3D supervision.
Neural Radiance Fields (NeRF)~\cite{mildenhall2021nerf} proposes to map each 3D point to a density value and a color taking viewing directions into consideration, enabling high-fidelity novel view synthesis using multi-view 2D supervision~\cite{zhang2020nerf++,zuo2023incremental}. 
However, extracting surfaces from NeRF's density field typically yields unsatisfactory results. 
A line of works hence replaces the density field with a signed distance or an occupancy field to improve reconstruction quality~\cite{yu2022monosdf,yariv2021volume,oechsle2021unisurf}. 
These methods focus on reconstructing the full scene and are not designed to decompose an arbitrary target object.

\subsection{Decomposed Neural Implicit Representations}
Decomposing neural implicit representations into distinct parts is useful for scene editing~\cite{yang2021learning} and object-level reconstruction~\cite{wu2022object}
Most methods depend on 2D annotations to guide the decomposition. 
SemanticNeRF~\cite{zhi2021place} introduces predicting a semantic logit at every 3D point, enabling scene classification while reconstruction. 
This idea is further explored for semantic segmentation~\cite{ranade2022ssdnerf,zhi2022ilabel}, panoptic segmentation~\cite{fu2022panoptic,liu2023instance}, and decomposed reconstruction~\cite{wu2022object,wu2023objectsdf++}.
These methods perform well with dense and accurate 2D annotations as supervision but require extensive human labeling, limiting their broader application.
Other approaches use category-level knowledge \cite{ranade2022ssdnerf,xie2021fig} or object motion \cite{martin2021nerf,wu2022d} to help decomposition. 
In contrast, we focus on reconstructing a prominent foreground object of arbitrary category in a cluttered scene without assuming it is dynamic.

\subsection{Self-supervised Learning in 3D Representations}
Recently, transformer architectures~\cite{dosovitskiy2020image,ranftl2021vision} and self-supervised learning methods have improved the generalization of image vision models~\cite{caron2021emerging,amir2022deep}.
Some methods decompose the neural implicit representations without 2D annotations or prior knowledge by utilizing 2D features from pre-trained networks~\cite{chen2024towards,yu2021unsupervised,fan2022nerf}.
NeRF-SOS~\cite{fan2022nerf} uses unsupervised 2D features from the visual transformer network DINO~\cite{caron2021emerging} to constraint segmentation and geometrical similarities between patches. 
DFF~\cite{kobayashi2022decomposing} exploits features from the visual transformer network DINO to supervise a proposed feature field with DINO feature maps. 
While we share the idea of using the 2D features from pre-trained methods as foreground-sensitive guidance, we argue that combining different types of DINO features yields a more stable result. 
We compare with DFF on varied datasets to validate our method.
AutoRecon~\cite{wang2023autorecon} aggregates DINO features to point clouds for coarse foreground segmentation but assumes a planar background for decomposition, making it difficult to handle complex environments.

\section{Methods}
Given a set of N images $\left\{\bI_1,...,\bI_N\right\}$ and the camera poses 
$\left\{\bxi_1,...,\bxi_N\right\}$, 
our goal is to learn decomposed object reconstruction that separates the background from the target object and vice versa. 

\subsection{Decomposed Neural Networks}

\textbf{Pipeline.} 
As shown in Fig.~\ref{fig:overview}, our method takes a 3D point position $\bx$ and a viewing direction $\bd$ as input, and then processes them with a background model and a foreground model, respectively. 
We follow~\cite{zhang2020nerf++} to build the background model, utilizing MLPs to obtain a density value $\sigma_{b}$ and a color vector $\bc_{b}$: $f_{\theta}: (\bx, \bd) \mapsto (\bc_b, \sigma_b).$

To reconstruct the surface of the foreground object, we follow~\cite{wang2021neus} to predict an SDF value and a color vector $\bc_{f}$, where the SDF value is transferred to a point opaque density $\rho_{f}$: $f_{\phi}: (\bx, \bd) \mapsto (\bc_f, \rho_f).$

A point-wise composition mechanism is then designed to blend the outputs of two models to get unified outputs $\bc$ and $\sigma$. 
Last, we use volume rendering~\cite{mildenhall2021nerf} to obtain pixel-wise colors $\bC$ and accumulated weight $W$. 
Thanks to the independent branch design, we can obtain the rendering results for the target object by rendering with the outputs of the foreground model.
Also, object surfaces could be extracted by applying marching cubes to the outputs of the foreground model.

\begin{figure}[tb]
\begin{center}
   \includegraphics[width=0.7\linewidth]{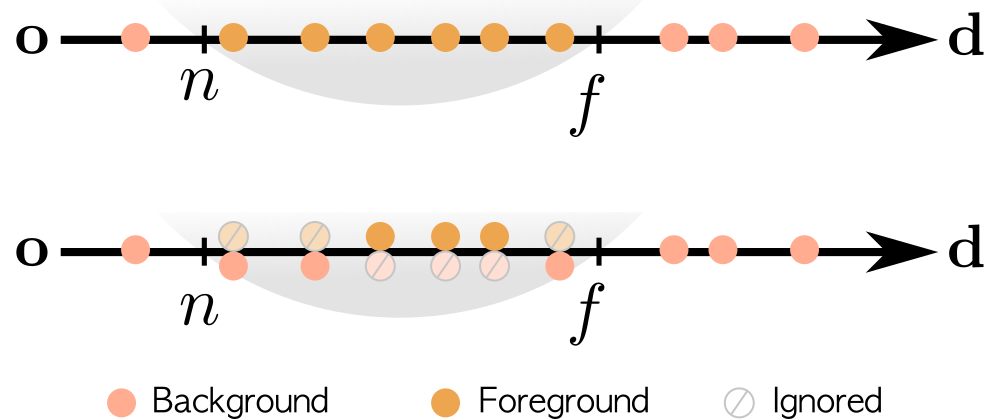}
\end{center}
    \vspace{-0.1cm}
   \caption{\textbf{Point-wise max composition.} 
   Unlike~\cite{wang2021neus}, we model points inside the unit sphere using both the foreground and background models. The values are combined via max composition, meaning the lower density is ignored.
   }
\label{fig:comp}
\vspace{-0.7cm}
\end{figure}

\textbf{Point-wise max composition.}
Following~\cite{zhang2020nerf++}, we adopt inverted sphere parameterization to partition the scene into an inner unit sphere and an outer volume.
As illustrated in Fig.~\ref{fig:comp}, the inner sphere is \textit{jointly} modeled with the foreground model $f_\theta$ and the background model $f_\phi$, whereas the rest is represented using the background model alone. This allows us to decompose the background region in the unit sphere from the foreground model.

For each point within the inner sphere, we adopt a point-wise composition to fuse the foreground and the background models. In our preliminary experiments, we experimentally observe that a simple max composition leads to better decomposition:
\begin{align}\label{eq:composition}
    \sigma &= \max(\sigma_{b},\rho_{f}) \\
    \bc &= \bc_{b}\mathds{1}_{\sigma_{b}>\rho_{f}}+\bc_{f}\mathds{1}_{\sigma_{b}<\rho_{f}}
\end{align}
where $\mathds{1}_{\star}\rightarrow \{0, 1\}$ refers to an indicator function.

\begin{table*}[t]\small
\renewcommand{\arraystretch}{1.1}
\centering
\setlength{\tabcolsep}{2.7mm}
\begin{tabular}{l|cc|cc|ccc|c} 
 \hline
 Dataset & DINO-Coseg~\cite{amir2022deep} & SAM~\cite{kirillov2023segment} & SNeRF+D~\cite{zhi2021place} & DFF~\cite{kobayashi2022decomposing} & NeuS~\cite{wang2021neus} & NeuS+D & NeuS+S & Ours \\
 \hline
 BlendedMVS & 0.772 & 0.577 & 0.745 & \slash  & 0.571 & 0.773 & 0.824 & \textbf{0.872} \\
 Tanks Temples & 0.890 & 0.603 & 0.916 & \slash  & 0.558 & 0.838 & 0.603 & \textbf{0.946} \\
 DTU & 0.900 & 0.936 & 0.887 & \slash  & 0.792 & 0.861 & 0.949 & \textbf{0.971} \\
 OnePose & 0.840 & 0.765 & 0.826 & 0.862  & 0.397 & 0.807 & 0.932 & \textbf{0.961} \\
 YCB Video  & 0.672 & 0.770 & 0.703 & 0.812 & 0.492 & 0.0.623 & 0.855 & \textbf{0.942} \\
 \hline 
\end{tabular}
\vspace{-0.2cm}
\caption{
    \textbf{Quantitative results} of foreground segmentation performance with mIoU metric. (+D) means using~\cite{amir2022deep} masks, while (+S) means using~\cite{kirillov2023segment} masks.
    }
\label{tab:seg}
\vspace{-0.35cm}
\end{table*}

\begin{table*}[htb]\small
\renewcommand{\arraystretch}{1.1}
\centering
\setlength{\tabcolsep}{2.1mm}
\begin{tabular}{l|ccccc|c|c|ccccccc} 
\hline
 Dataset & \multicolumn{5}{c|}{BlendedMVS} & T\&T & DTU & \multicolumn{7}{c}{YCB Video} \\
 \hline
SceneID & 1 & 2 & 3 & 4 & Mean & 1 & 1 & 1 & 2 & 3 & 4 & 5 & 6 & Mean \\
\hline
NeuS & 0.073 & 0.161 & 1.075 & 0.240 & 0.387 & 0.105 & 0.348 & 0.120 & 0.098 & 0.059 & 0.326 & 0.085 & 0.623  & 0.219  \\
SNeRF+D & 0.093 & 0.129 & 0.336 & 0.192 & 0.188 & 0.079 & 0.349 & 0.133 & 0.114 & \textbf{0.036} & 0.319 & 0.069 & 0.489 & 0.193  \\
NeuS+D  & 0.140 & 0.122 & 0.339 & 0.156 & 0.189 & 0.086 & 0.250 & 0.153 & 0.170 & 0.082 & 0.302 & \textbf{0.057} & 0.352 & 0.188  \\
NeuS+S  & 0.072 & 0.117 & 0.451 & \textbf{0.108} & 0.187 & 0.081 & 0.191 & 0.135 & 0.131 & 0.041 & 0.334 & 0.063 & 0.355 & 0.177  \\
\hline
Ours  & \textbf{0.068} & \textbf{0.090} & \textbf{0.170} & 0.125 & \textbf{0.113} & \textbf{0.054} & \textbf{0.155} & \textbf{0.111} & \textbf{0.086} & 0.037 & \textbf{0.298} & 0.064 & \textbf{0.336} & \textbf{0.155}  \\
 \hline 
\end{tabular}
\vspace{-0.2cm} 
\caption{
    \textbf{Quantitative results} of reconstruction performance. We report CD to measure the reconstruction quality.
    }
\label{tab:recon}
\vspace{-0.7cm}
\end{table*}

\subsection{Leveraging 2D Self-Supervised Features}\label{sec:dino}
Decomposing the foreground and the background without manually labeling is difficult. 
Existing methods often use pre-trained segmentation models or self-supervised models to aid in this process, employing either binary masks or fine-grained features. 
Similarly, we advocate for using self-supervised 2D features for decomposition. However, to the best of our knowledge, we are the first to transfer self-supervised 2D features into masks of varying granularities for supervision, including median-grained and coarse-grained masks.

\begin{figure}[tb]
\begin{center}
   \includegraphics[width=0.95\linewidth]{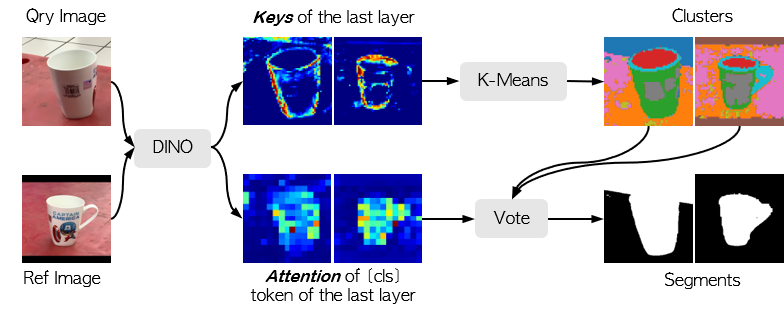}
\end{center}
\vspace{-0.3cm}
   \caption{\textbf{Median-grained and coarse-grained masks} obtained from self-supervised networks.}
\label{fig:dino}
\vspace{-0.5cm}
\end{figure}

\textbf{Median-grained mask supervision.}
We extract median-grained masks from self-supervised 2D features to facilitate decomposed reconstruction.
Fig.~\ref{fig:dino} shows the feature generation process following~\cite{amir2022deep}.
Each input image is paired with the reference image.
Their key descriptors and attention maps of the [CLS] token of the last transformer layer are extracted by DINO~\cite{caron2021emerging}. 
K-Means is then applied to cluster features of the key descriptors from both images, representing the common feature clusters appearing in both views.
Our key observation is that median-grained masks are less prone to error, i.e., the same cluster often belongs to the same foreground or background class. 
Thus, we design a cluster mask loss to ensure that the pixels in the same cluster fall into the same class.

\begin{figure}[tb]
\begin{center}
\includegraphics[width=0.95\linewidth]{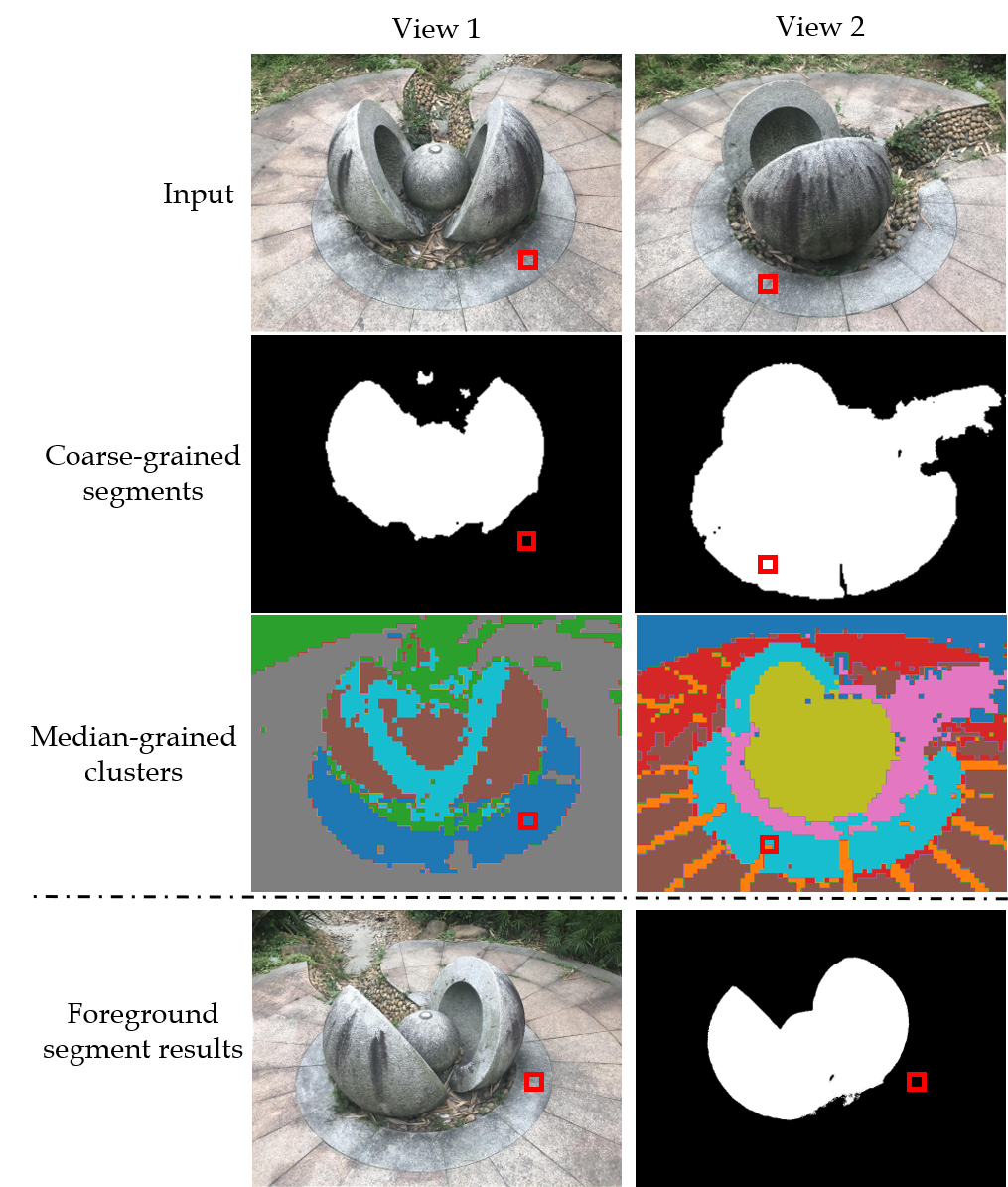}
\end{center}
\vspace{-0.3cm}
   \caption{
   \textbf{Coarse- and median-grained features} of the same scene in two different views.
   }
\label{fig:noisy}
\vspace{-0.72cm}
\end{figure}

Given a set of M sampled rays, we retrieve their cluster values 
$\left\{l_i \vert l_i \in \cS\right\}_{i=1}^{M}$, where $\cS=\left\{s_j\right\}_{j=1}^K$ is the cluster index set. 
We group sampled rays of the same cluster index and compute the average foreground weight of each cluster index $s_j$:
\begin{equation}
g_{s_j} = E(\{W_{f_{i}} \vert l_i = s_j\})    
\end{equation}
where $E(\cdot)$ is an average operation and $W_f$ is the foreground weight at one sampled ray.
Next, we use the binary cross entropy loss to push the average foreground weight of each cluster to either $0$ or $1$:
\begin{equation}\label{eq:clusterloss}
\cL_{cls_{s_j}}=-(g_{s_j}\log g_{s_j} + (1-g_{s_j})\log (1-g_{s_j}))
\end{equation}
We sum up the losses of every cluster to get the final cluster mask loss 
\begin{equation}\label{eq:cls}
\cL_{cls}=\sum_{j \in K}\cL_{cls_{s_j}}
\end{equation}
As illustrated in Fig.~\ref{fig:overview}, $\cL_{cls}$ is applied to each cluster. Hence it is cluster-level supervision instead of pixel-level.

\begin{figure*}[tb]
\begin{center}
\includegraphics[width=0.95\linewidth]{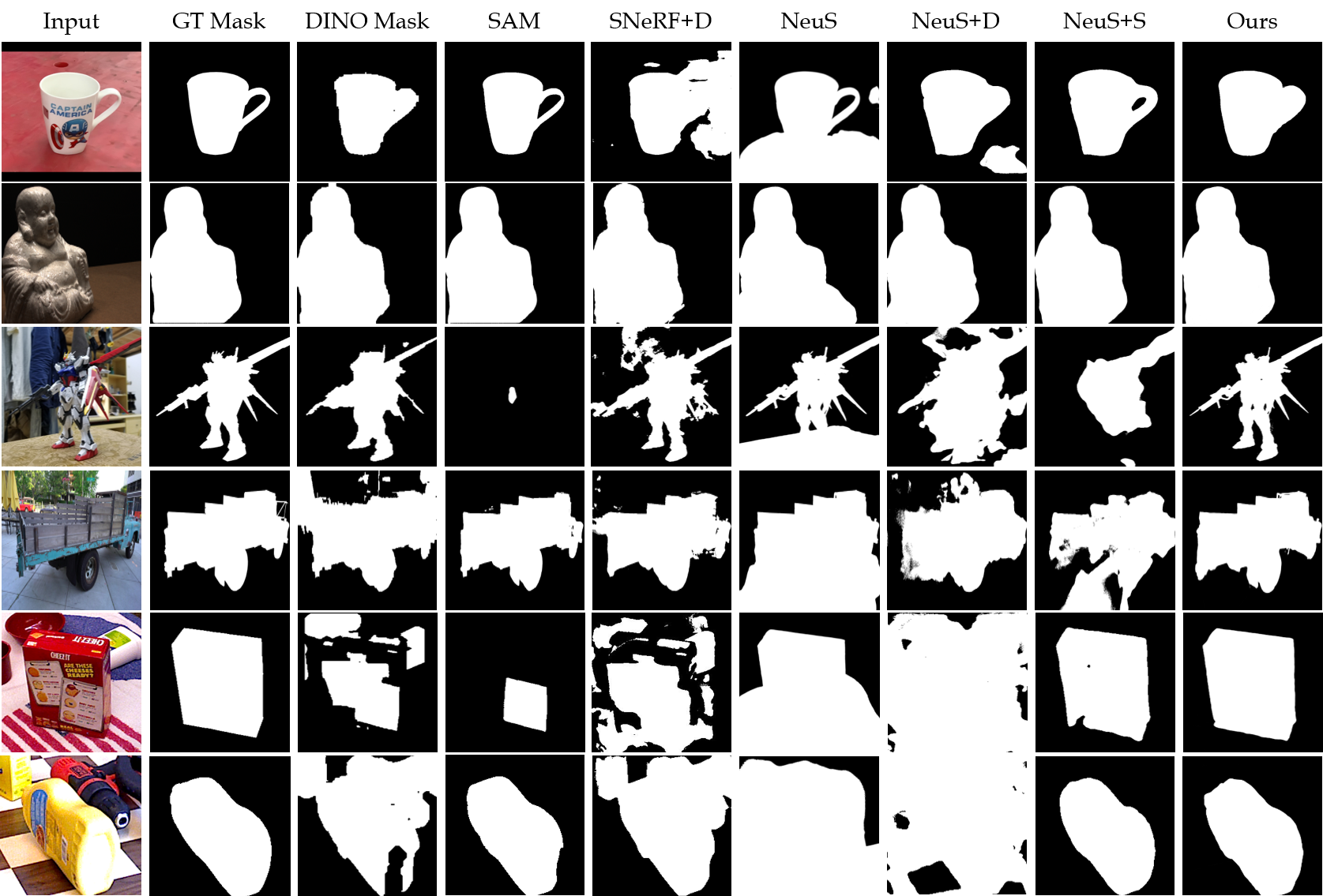}
\end{center}
\vspace{-0.3cm}
   \caption{
   \textbf{Qualitative foreground segmentation results} in multiple datasets.
   }
\label{fig:seg}
\vspace{-0.6cm} 
\end{figure*}

\begin{figure}[tb]
\begin{center}
\includegraphics[width=0.95\linewidth]{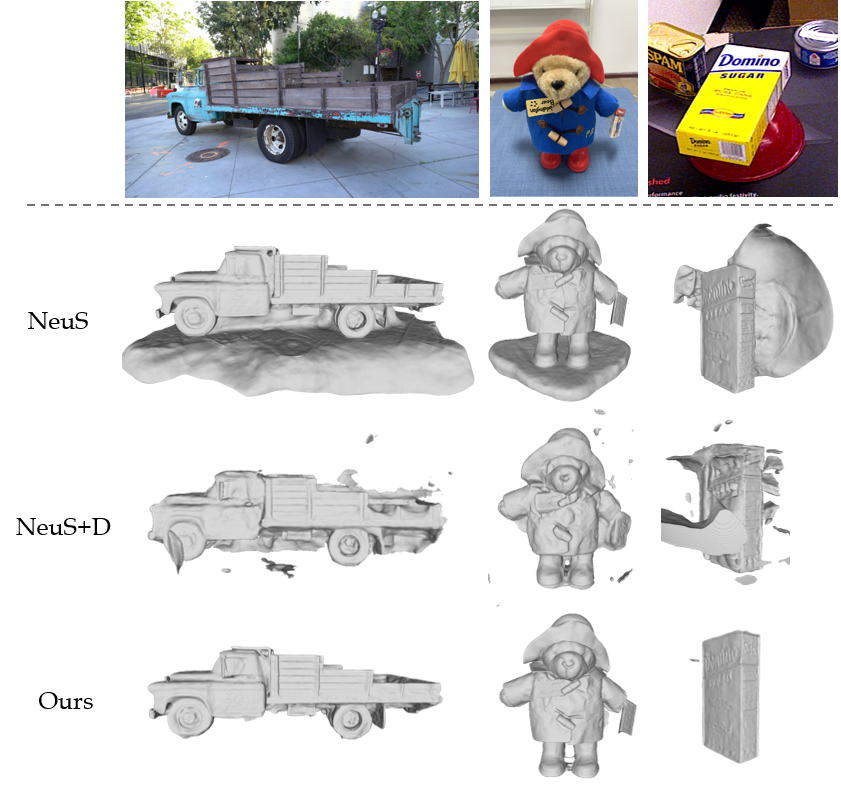}
\end{center}
\vspace{-0.45cm}
   \caption{\textbf{Qualitative reconstruction results} in Tanks and Temples, YCB Video, and BlendedMVS datasets. }
\label{fig:recon}
\vspace{-0.6cm}
\end{figure}

\textbf{Coarse-grained mask supervision.}
By utilizing the cluster mask loss, rays belonging to the same feature cluster are encouraged to be classified into the same class, aiding in a clean decomposition.
However, this per-cluster loss alone does not specify which region should be in the foreground,
resulting in many possible solutions that can drive the loss close to zero, which is unfavorable for initial network convergence. 
Therefore, we propose integrating coarse-grained binary masks to guide early-stage training.

As shown in Fig.~\ref{fig:dino}, we follow \cite{amir2022deep} to use a simple voting procedure to select clusters that are salient and common to both images as our binary mask, indicating a separation between the foreground and the background.
We deploy the coarse-grained binary masks to supervise the weights map rendered from the foreground model by the binary cross entropy loss $\cL_{seg}$. This per-pixel loss provides a clear preference of which region should be classified as foreground.

\textbf{Analysis of coarse- and fine-grained masks.}
We further analyze how the two proposed features work under difficult situations.
Fig.~\ref{fig:noisy} shows an example of the obtained 2D self-supervised features of the same scene in two views.
The foreground stone and the background floor tiles appear similar. 
Due to the lack of inter-frame awareness in the 2D self-supervised network's attention activation, they are classified into different classes in the coarse-grained segment features in the 1st view, while both are classified as foreground in the 2nd viewpoint (see red box).
This leads to inconsistent convergence directions when training our network with different observations.
In contrast, median-grained features utilize richer information to provide a \emph{per-cluster} form of supervision.
Note, that such median-grained masks can correctly group foreground and background into the same clusters.
Under our cluster mask loss in Eq.~(\ref{eq:cls}),
each pixel must align with the results of most other pixels within the cluster during convergence. 
Therefore, if the cluster is correctly classified in most observations, it ensures that all pixels within the cluster are correctly classified,
enhancing the robustness compared to using the coarse-grained mask alone.

\subsection{Training}
\textbf{Losses.}
Following \cite{mildenhall2021nerf}, we minimize the reconstruction error between the predicted color $\hat{\bC}(\br)$ and the ground truth color $\bC(\br)$
\begin{equation}\label{eq:colorloss}
\cL_{color}=\Vert\hat{\bC}(\br)-\bC(\br)\Vert_2^{2}
\end{equation}

Since we represent the foreground object as implicit surfaces, an Eikonal loss $\cL_{eik}$~\cite{gropp2020implicit} is utilized to regularize the predicted SDF values.

To prevent the background model from representing all of the points inside the sphere, we add a loss to punish the foreground model when its weighted alpha sum approaches zero
\begin{equation}\label{eq:fgreg}
\cL_{fg}= -log(\frac{\beta}{B}\sum_{i}(\mathds{1}_{W_{f_{i}}>W_{th}}))
\end{equation}
where $W_{f}$ is the weighted alpha sum from the foreground model, $W_{th}$ is a threshold value, $\beta$ is a scale hyper-parameter, and $B$ is the number of rays.

Combing with the median-grained and coarse-grained mask supervision, our full loss function is as follows:
\begin{equation}
\cL=\gamma_1 \cL_{color}+\gamma_2 \cL_{eik}+\gamma_3 \cL_{fg}+\gamma_4 \cL_{seg}+\gamma_5 \cL_{cls}
\end{equation}

\textbf{Training strategy.}
We experimentally observe that the training strategy is important for achieving stable decomposition. Specifically, we first train the model from scratch using $\cL_{color}$, $\cL_{eik}$, $\cL_{fg}$ and $\cL_{seg}$, as we observe that the coarse-grained mask supervision $\cL_{seg}$ provides a clear indicator of foreground/background separation, leading to a reasonable initial decomposition. Next, we fine-tune the full model with all losses, where $\cL_{cls}$ plays the main role in decomposing the foreground.

\begin{table}\small
\centering
\setlength{\tabcolsep}{2.7mm}
\begin{tabular}{l|cccc} 
 \hline
 & s0  & s1 & s2 & s3 \\
 \hline
binary mask loss (sc) &  & \checkmark &  & \checkmark \\ 
cluster mask loss (sc) &   &  & \checkmark &   \\
 cluster mask loss (ft)  &  &  & & \checkmark \\
 \hline
 scene 1 & 0.562 & 0.763 & 0.967 &  0.966  \\
scene 2 & 0.642 & 0.705 & 0.319 &  0.958 \\
 \hline
\end{tabular}
\vspace{-0.2cm}
\caption{\textbf{Ablation study} on training strategies. Here mIoU is reported to indicate the decomposition performance. }
\label{tab:ab1}
\vspace{-0.5cm}
\end{table}

\begin{figure}[t]
\begin{center}
\includegraphics[width=0.98\linewidth]{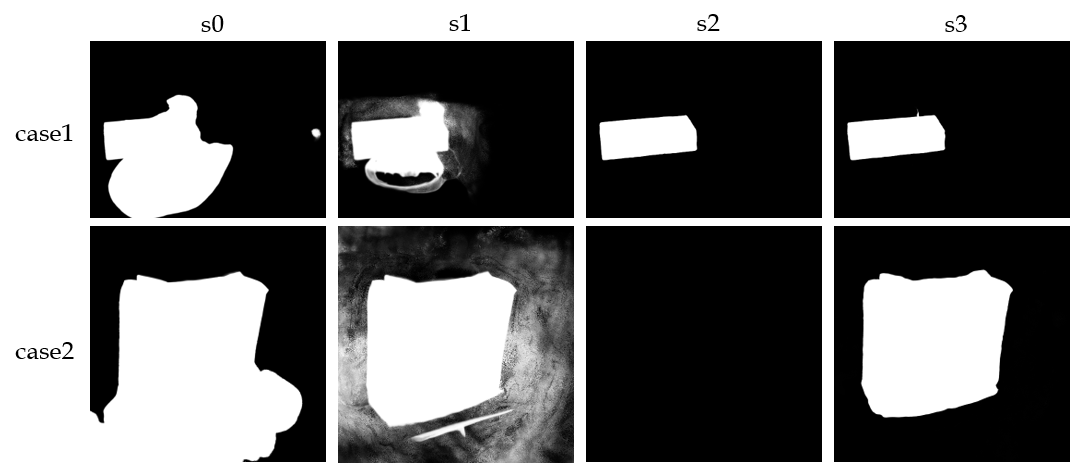}
\end{center}
\vspace{-0.3cm}
   \caption{
   Visualization of ablation study on training strategy.
   }
\label{fig:add-ablation}
\vspace{-0.7cm}
\end{figure}

\section{Experiments}

We evaluate our approach to answer two questions: 
i) can our method reconstruct compact object geometries and obtain neat segmentation results when synthesizing novel views? 
ii) does our design in utilizing 2D unsupervised features enhance performance?

\subsection{Datasets and metrics}
\textbf{Datasets.} 
We conduct experiments on 
5
different real world datasets to validate our method. 
\textbf{BlendedMVS} (BMVS)~\cite{yao2020blendedmvs}, 
\textbf{DTU}~\cite{jensen2014large}
, and \textbf{Tanks and Temples} (T\&T)~\cite{knapitsch2017tanks} are commonly used in 3D reconstruction. 
We manually label the test images and remove the background in the provided meshes as ground truths.
\textbf{OnePose}~\cite{sun2022onepose} is an object-centered dataset, containing common domestic products placed on different platforms and observed from top-down views. 
\textbf{YCB Video}~\cite{xiang2017posecnn} is a more challenging dataset, where objects could be placed on a non-planar background.
We follow~\cite{zhang2020nerf++} to place a sphere to enclose the target object. 
\footnote{More implementation and dataset details are presented in the Appendix: https://arxiv.org/abs/2310.11092\label{ft1}}

\textbf{Metrics.}
We employ the following metrics for evaluation: \textbf{mIoU} (mean Intersection-over-Union) to evaluate the decomposed segmentation results, and \textbf{CD} (Chamfer Distance)
to measure the quality of reconstructed 3D geometry.

\begin{table}[t]\small
\centering
\setlength{\tabcolsep}{3.5mm}
\begin{tabular}{l|ccccc} 
 \hline
offset & 0 & 0 & 0 &  5$\%$ & 10$\%$ \\
scale & 1.0 & 0.9 & 1.2 & 1.0 & 1.0 \\
\hline
mIoU & 0.920 & 0.924 & 0.920 & 0.918 & 0.922 \\
CD & 0.227 & 0.244 & 0.206 & 0.230 & 0.239 \\
 \hline
\end{tabular}
\vspace{-0.15cm}
\caption{
    \textbf{Ablation study} on sphere parameters.
    }
\label{tab:ab-shift}
\vspace{-0.3cm}
\end{table}

\begin{table}[t]\small
\centering
\setlength{\tabcolsep}{3.5mm}
\begin{tabular}{l|ccccc} 
 \hline
$\mathcal{L}_{fg}$ & 1 & 2 & 3 & 4 & Mean \\
 \hline
 & 0.776 & 0.872 & 0.965 & 0.640 & 0.813 \\
 \checkmark & 0.784 & 0.879 & 0.964 & 0.862 & 0.872 \\
 \hline
\end{tabular}
\vspace{-0.15cm}
\caption{
    \textbf{Ablation study} on foreground regularization loss $\mathcal{L}_{fg}$ with the metric of mIoU.
    }
\label{tab:ab-fgloss}
\vspace{-0.65cm}
\end{table}

\begin{table}[t]\small
\centering
\setlength{\tabcolsep}{3mm}
\begin{tabular}{c|ccccc} 
 \hline
seg. noise & 0 & 5$\%$ & 10$\%$ & 0 & 0 \\
cls. noise & 0 & 0 & 0 & 5$\%$ & 10$\%$ \\
\hline 
mIoU & 0.947 & 0.946 & 0.942 & 0.951 & 0.954 \\
 \hline
\end{tabular}
\vspace{-0.1cm}
\caption{
    \textbf{Ablation study} on foreground decomposition results towards feature noises.
    }
\label{tab:ab-noise}
\vspace{-0.4cm}
\end{table}

\subsection{Implementation details}
For each object, the RGBs are paired with the first image to obtain DINO features.
For YCB Video scenes, we use the bounding rectangle of the projected sphere on the 2D plane as a bounding box and extract DINO features from within the box. 
For other scenes, since they are already object-centered, we use the whole image for DINO.
We train each object for 200k iterations with color loss, eikonal loss, foreground regularization loss, and binary mask loss from scratch, and finetune with cluster mask loss from 150k. 
It takes about 4 hours to reconstruct one object on a single GTX 3090Ti GPU, requiring around 7.3GB of GPU memory.
We use the official implementation of~\cite{amir2022deep} to obtain both the coarse-grained and median-grained features without any parameter-tuning for all experiments.
\footref{ft1}

\subsection{Comparison to the State of the Art}

\textbf{Foreground Segmentation. }
We report the segmentation results in all datasets in Tab.~\ref{tab:seg}. 
Our method outperforms other baselines in most scenes with mIoU metrics.
For the 2D self-supervised learning methods (DINO-Coseg and SAM), they use image feature attentions to separate foregrounds from backgrounds but often mistakenly respond to textures or shadows.
As for the 3D reconstruction methods, the performance of NeuS shows the results without background modeling inside the sphere or utilization 2D self-supervised features, serving as a naive baseline of our method.
SNeRF+D and NeuS+S, relying on 2D self-supervised masks for supervision, show similar results to their 2D counterparts due to strong mask-fitting abilities.
DFF learns a feature field to aggregate self-supervised features from multi-frames but neglects the cluster similarities and requires a query feature from DINO for post-segmentation.
Due to the larger image sizes on the BMVS and TT datasets, we have to use larger strides to extract dino features. 
However, training DFF with such coarse dino features yields less satisfactory results, so we excluded these results.

Fig.~\ref{fig:seg} shows the qualitative results in OnePose, BlendedMVS, and YCB Video datasets. Our method decomposes the target object from the background neatly in most cases.

\textbf{Foreground Reconstruction.}
To validate the object-level reconstruction and decomposition ability in different objects and environments, we compare our method in 4 datasets with 3D reconstruction baselines. 
Since the OnePose dataset doesn't provide ground truth meshes, the comparison is spared.

As shown in Table.~\ref{tab:recon}, our method performs best in most scenes, demonstrating superior reconstruction accuracy and completeness.
NeuS, serving as a naive baseline, highlights the difference between the object mesh and the background-contained mesh. 
The NeRF-based method SNeRF+D shows inferior reconstruction results.
NeuS+S takes advantage of the strong surface implicit representation and state-of-the-art unsupervised segmentation masks, sometimes performs better but is unstable, showing sensitivity to supervision masks.
Fig.~\ref{fig:recon} displays the qualitative results of our method and the baselines in varied datasets. 
Notably, the result in YCB Video (column 3) demonstrates our ability to decompose the target object from non-planar backgrounds.

\subsection{Ablation studies}

\textbf{Effectiveness in training strategy.}
To validate the effectiveness of our training strategy in utilizing two kinds of features, we conduct ablation studies on two different scenes.
Table.~\ref{tab:ab1} demonstrates the performance difference: (sc) indicates training from scratch, and (ft) refers to fine-tuning. 
Fig.~\ref{fig:add-ablation} shows the qualitative results.

\begin{figure}[t]
\begin{center}
\includegraphics[width=0.95\linewidth]{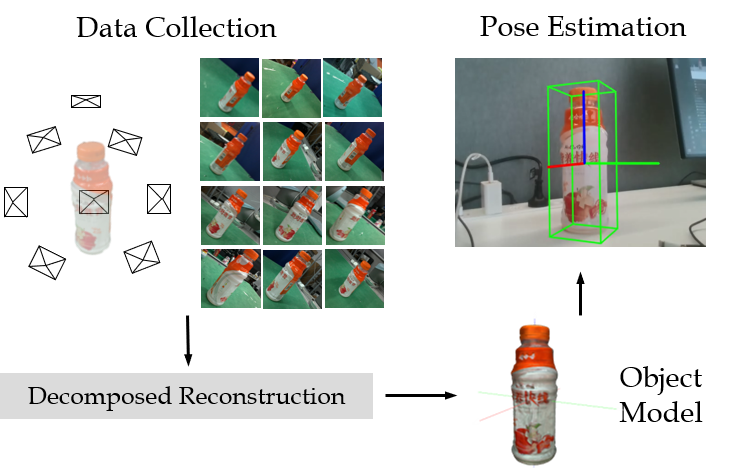}
\end{center}
\vspace{-0.3cm}
   \caption{
   \textbf{Real-world experiment process.}
   }
\label{fig:real-set}
\vspace{-0.7cm}
\end{figure}

\begin{figure*}
\begin{center}
\includegraphics[width=0.98\linewidth]{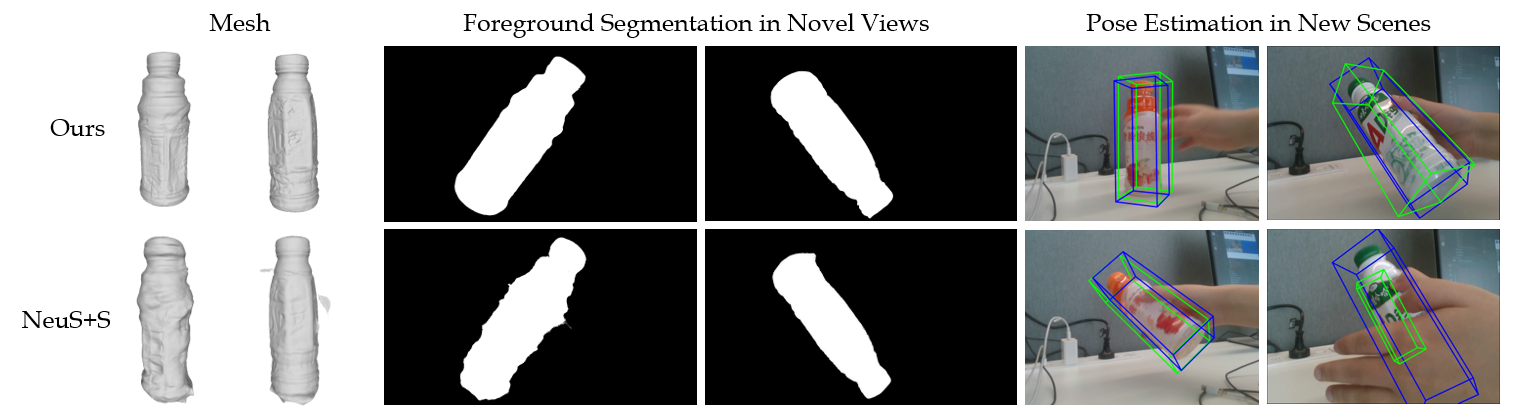}
\end{center}
\vspace{-0.3cm}
   \caption{
   \textbf{Qualitative results on real-world experiments:} reconstruction, foreground segmentation, and pose estimation.
   }
\label{fig:real-res}
\vspace{-0.5cm}
\end{figure*}

\begin{table}[t]\small
\centering
\setlength{\tabcolsep}{4mm}
\begin{tabular}{l|ccc} 
 \hline
 & Bottle1 & Bottle2 & Mean \\
 \hline
NeuS+S & 0.903 & 0.948 & 0.926  \\
Ours & 0.972 & 0.984 & 0.978  \\
 \hline
\end{tabular}
\vspace{-0.15cm}
\caption{
    \textbf{Quantitative results on real-world experiments} of foreground segmentation with the metric of mIoU. 
    }
\label{tab:real-quan}
\vspace{-0.95cm}
\end{table}

As shown, the network struggles to decompose when neither binary mask loss nor cluster mask loss is used (s0). 
Providing binary mask loss (s1) improves results, but artifacts remain due to segment noise.
Using only cluster mask loss (s2) results in one object decomposing well while failing on another, indicating that higher granularity features can misguide the network.
Training with binary mask loss from scratch and then fine-tuning with cluster mask loss (s3) yields the most accurate and stable results. Therefore, we adopt the s3 strategy for all experiments in this paper.

\textbf{Sensitivities to sphere parameterization.}
We follow \cite{wang2021neus} to use the inverted sphere parameterization, where the sphere determines the potential object range. 
To investigate the influence of sphere parameter perturbations, we conduct ablation studies by scaling the sphere radius and introducing offsets to the sphere center. 
The offset magnitude is given as a percentage of the sphere’s radius. 
As shown in Table.~\ref{tab:ab-shift}, variations in sphere parameters within certain limits do not affect our results significantly, highlighting the robustness of our method's decomposition ability.

\textbf{Ablation studies on foreground regularization.}
We validate the effectiveness of the proposed foreground regularization loss $\mathcal{L}_{fg}$ in the BlendedMVS dataset.
As shown in Table.~\ref{tab:ab-fgloss}, omitting $\mathcal{L}_{fg}$ has little effect on three scenes but significantly decreases performance in one scene. 
This indicates the importance of the error term in enhancing the method's robustness.

\textbf{Sensitivities to self-supervised feature qualities.}
We report how the noises in the obtained features affect our results.
We add random noises to the coarse-grained features and median-trained features to assess its effect on decomposing results. 
As shown in Table.~\ref{tab:ab-noise}, with feature contamination, our segmentation performance remains strong, demonstrating the robustness of our approach.

\subsection{Real-world Experiments}
We conduct real-world experiments to validate our performance and show how the decomposed reconstruction quality affects downstream tasks.
Here, we consider zero-shot object pose estimation as one representative downstream task.
Fig.~\ref{fig:real-set} shows our experiment process.
First, we collect data by placing target objects on a platform and capturing about 60 multi-view images per object with a moving camera. 
70\% of these images are used for reconstruction training, with the remainder for novel view testing.
Next, we feed the data to two reconstruction methods for comparison: our proposed method and the NeuS+S baseline.
All parameters and settings are consistent with our public dataset experiments.
Last, we utilize the obtained models in new test scenes for a downstream object pose estimation task using FoundationPose~\cite{wen2024foundationpose}.
This task involves estimating the object's pose based on the observations and the template image rendered from an object model. 
While FoundationPose uses CAD models, we replace them with our decomposed object model.
The quality of the object model directly affects the estimation results, making it a suitable task to indicate why clean decomposed reconstruction is essential.

\textbf{Results on decomposed reconstruction.}
Since we lack ground truth models of the real-world objects, we compare the results of decomposed foreground segmentation in novel views with human-annotated ground truth masks.
Table.~\ref{tab:real-quan} displays the quantitative results, while Fig.~\ref{fig:real-res} shows the qualitative results of our method and the baseline.
As shown, the baseline method can decompose both the bottles from the background but fails to achieve complete modeling. 
In contrast, our method achieves consistent decomposed modeling results.

\textbf{Results on downstream task.}
We directly apply ~\cite{wen2024foundationpose} to two test scenes using models from both our method and the baseline.
Lacking ground truth meshes for the target objects, we compare their estimation performance qualitatively.
Fig.~\ref{fig:real-res} displays the estimation results: the blue bounding box shows results using our model, while the green box shows the results using the baseline method.
The baseline method's incomplete decomposition can lead to larger estimation errors in occlusion cases, as seen in the lower right image.
More qualitative results are available in the supplementary video.

\section{Conclusion}
In this work, we propose a Decomposed Object Reconstruction method (DORec) based on neural implicit representation, which learns to decompose the target object from the background while reconstructing.
Our method transfers 2D self-supervised features into masks of two levels of granularity to supervise the decomposition, including a binary mask to indicate the foreground regions and a K-cluster mask to indicate the semantically similar regions.
Extensive experimental results on real datasets demonstrate the strong ability of our method in both segmentation, object-level reconstruction, and downstream tasks.
A limitation is the lengthy training time, which we aim to reduce in the future using advanced network architectures or training strategies.

\bibliographystyle{IEEEtran}

\bibliography{root}

\clearpage
\newpage
\begin{appendix}\label{appendix}

\begin{figure*}[t]
\begin{center}
\includegraphics[width=0.95\linewidth]{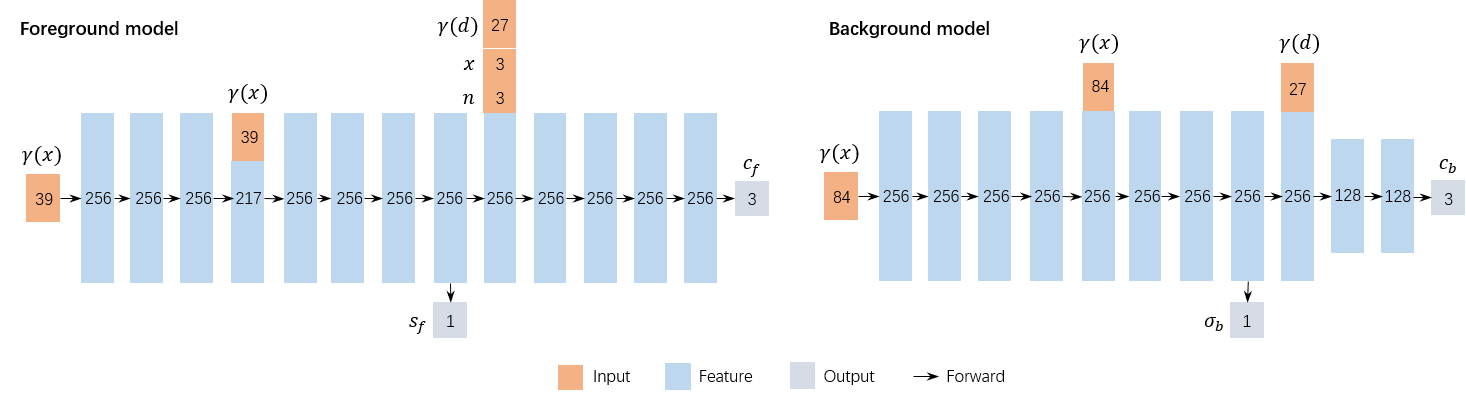}
\end{center}
   \caption{Network Architecture of DORec.}
\label{fig:detailed_network}
\end{figure*}

\subsection{Preliminaries}
\textbf{Surface-based Implicit Representation.}
We use neural surface-based implicit representation as our foreground model for extracting the object's surface.

Let $\br(t)=\bo+t\bd$ be a camera ray emitted from the focal principle $\bo$ of a camera with a unit-norm viewing direction $\bd$, a set of 3D points $\bx$ are sampled on the ray within a bounded 3D volume.
Given $(\bx,\bd)$ as inputs, we follow~\cite{wang2021neus} to project each input to a color value $\bc(t)$ and an SDF value $s(t)$ with MLPs. Then, the opaque density $\rho(t)$ is computed as
\begin{equation}\label{eq:sr}
\rho(t)=\max \left(\frac{-\frac{d\Phi}{dt}(s(t))}{\Phi(s(t))}, 0\right).
\end{equation}

The sigmoid function $\Phi(s(t))=(1+e^{-\beta s(t)})^{-1}$ is the integral of the logistic density distribution of the SDF value. To obtain pixel color $\bC(\br)$ and accumulated weight $W(\br)$, volume rendering is adopted:
\begin{equation}\label{eq:vr}
\bC(\br)=\int_{n}^{f}T(t)\alpha(t)\bc(t)dt, \quad  W(\br)=\int_{n}^{f}T(t)\alpha(t)dt,
\end{equation}
where $n$ and $f$ refer to the near and far points for integration, $T(t)$ is the accumulated transmittance defined as~\cite{mildenhall2021nerf}, and $\alpha(t)$ is the alpha value derived from $\rho(t)$~\cite{wang2021neus}.

\textbf{Self-supervised Visual Transformer.}
Vision Transformers (ViT)~\cite{dosovitskiy2020image} split an input image into a sequence of fixed-size patches with position embedding, and feed the sequence to a Transformer encoder. To perform image classification, an extra learnable embedding "classification token", noted as [CLS], is added to the sequence. 
Perceiving self-attention and cross-attention with the patches in the encoder, the output of the [CLS] token goes through an MLP layer to obtain the final classification probabilities.
DINO~\cite{caron2021emerging} deploys the ViT in a self-supervised training manner, providing foreground-sensitive features without a predefined label set. 

\subsection{Implementation Details of DORec}
\textbf{Network Architecture.} Fig.~\ref{fig:detailed_network} shows the detailed architecture of our DORec model. We adopt the same network architecture in all experiments. We follow~\cite{wang2021neus} to build our foreground model and background model. 

The foreground model takes as input the 3D location $\bx$ to predict the signed distance field value $\bs_{f}$. Then it calculates the point-wise gradients $\bn=d\bs/d\bx$, and concatenates them with the feature vector, the viewing direction $\bd$, and the location $\bx$ to predict the color field value $\bc_f$. The background model takes as input the 3D location $\bx$ to predict the density field value $\sigma_{b}$. Then it combines the viewing direction $\bd$ to get the color field value $\bc_b$.

Following NeRF~\cite{mildenhall2021nerf}, both $\bx$ and $\bd$ are mapped to a higher dimensional space using a positional encoding (PE):
\begin{equation}
\begin{split}
\gamma(\bp) = (sin(2^{0}\pi \bp),cos(2^{0}\pi \bp),...,\\ 
    sin(2^{L-1}\pi \bp),cos(2^{L-1}\pi \bp)) 
\end{split}
\end{equation}
In the foreground model, we set $L=6$ for $\gamma(\bx)$ and $L=4$ for $\gamma(\bd)$. In the background model, we set $L=10$ for $\gamma(\bx)$ and $L=4$ for $\gamma(\bd)$.

\textbf{Sampling Strategy.} We normalize the scene, placing the sphere at the center with a radius of 1. 
We sample 128 points for the foreground model between the range of [-1,1] with the hierarchical sampling strategy proposed by~\cite{wang2021neus}. 
Specifically, we first uniformly sample 64 points along the ray, then iteratively conduct importance sampling for 4 times. In each iteration, we additionally sample 16 points. 
Then we sample extra 64 points outside the normalized sphere and normalize them following~\cite{zhang2020nerf++}. We concatenate the outside points (64) with the inside points (128) together to feed the background model. When extracting mesh, we evenly sample $512\times512\times512$ points inside a cube with the radius of the sphere and use their predicted SDF value for marching cubes~\cite{lorensen1987marching}.

\textbf{Training Strategy.} We randomly select one image from the trainset and sample 512 rays in each iteration. We use the Adam optimizer to train our networks. The learning rate is first linearly warmed up from 0 to $5\times10^{-4}$ in the first 5k iterations, and then controlled by the cosine decay schedule to the minimum learning rate of $2.5\times10^{-5}$. We train each model for 150k iterations with the losses $\cL_{color}$, $\cL_{eik}$, $\cL_{fg}$, and $\cL_{seg}$ mentioned in the paper, then add $\cL_{clu}$ to finetune for another 50k iterations.
The loss weights ${\gamma}_{i}$ are set to 1.0, 0.1, 0.01, 0.1, and 0.1, respectively. 
Every model is trained on a single GTX 3090Ti GPU. 

\textbf{Applying DINO-Coseg.} As described in the paper, we propose to transfer self-supervised 2D features to masks of different granularities.
We use the official implementation of~\cite{amir2022deep}.
We adopt the ViT-S/8 network with a stride of 4 for the first convolutional layer and set the maximum cluster number as 16.
The feature-obtaining process is conducted before model training. It takes about 2.5s to process one image in an NVIDIA 3090Ti GPU.
We refer our readers to~\cite{amir2022deep} for more feature extraction network details.

\subsection{Implementation Details of Baseline Methods}

\textbf{DINO-Coseg:} For DINO-Coseg~\cite{amir2022deep}, we use the same coarse-grained binary masks as our method use for fair comparison.

\textbf{SAM:} For SAM~\cite{kirillov2023segment}, we use the official implementation provided by the authors. We adopt the public trained model and set the prompt point per side as 1 to get a binary mask. In BlendedMVS, OnePose, and Tanks and Temples datasets, we use the whole image as a prompt to obtain a foreground and background mask. While in the non-object-centered dataset YCB Video, we use the same 2D bounding boxes as we use as prompts.

\textbf{NeuS:} For NeuS~\cite{wang2021neus}, we use the official implementation provided by the authors, except that we mask out the black edges in the images in OnePose dataset when calculating color loss.
For NeuS+D and NeuS+S, we use the DINO-Coseg masks and SAM masks respectively.

\textbf{SemanticNeRF:} For SemanticNeRF~\cite{zhi2021place}, we also use the official implementation provided by the authors. We use the binary we obtained from DINO-Coseg as supervision in training SemanticNeRF. In extracting mesh, we also uniformly sample $512\times512\times512$ points in the cubic encasing the sphere to predict their densities and semantics, then use argmax to choose those points belonging to foreground, and adopt marching cubes to generate mesh.

\textbf{DFF:} For DFF~\cite{kobayashi2022decomposing}, we use the official implementation provided by the authors. Since the code for DINO feature extraction is not presented, we follow the introductions in their paper~\cite{kobayashi2022decomposing} to deploy~\cite{amir2022deep} for feature extracting.
Also, DFF requires a reference patch as input to extract the interested region. We use the GT object mask to choose the object region in the first input image, then calculate the mean feature vector in the region as the query feature to select the target object in every input image.
We consider annotating a GT object mask at one view of a similar cost of annotating a sphere, and thus we do not apply a sphere projection to post-process the DFF results.

\subsection{Datasets}\label{sec2}

\textbf{OnePose Dataset:}
OnePose dataset~\cite{sun2022onepose} is firstly proposed for evaluating one-shot object pose estimation methods. We choose this dataset to evaluate our method because it contains various common products in different environments with real lighting. 
We downsample the original image sequences with a ratio of 10 to get training images, and randomly sample from the rest to get testing images.
Tab.~\ref{tab:onepose} shows the specific cases, sequences and numbers of images we use for training and testing.
We use the official implementation provided by the authors to obtain the poses and a sparse pointcloud of each object. Since no ground truth masks are provided in the dataset, we manually label all the testing images.

\begin{table}\small
\renewcommand{\arraystretch}{1.1}
\centering
\setlength{\tabcolsep}{2mm}
\begin{tabular}{c|c|c|c|c} 
 \hline
Scene & Case & Sequence & Train\# & Test\#  \\
 \hline
1& 0413-juliecookies-box & 3 & 63 & 17 \\
2& 0414-babydiapers-others & 1 & 73 & 18 \\
3& 0415-captaincup-others & 3 & 63 & 17 \\ 
 \hline
\end{tabular}
\caption{Sequences and images used in OnePose dataset.}
\label{tab:onepose}
\end{table}

\begin{table}[ht]\small
\renewcommand{\arraystretch}{1.1}
\centering
\setlength{\tabcolsep}{2.5mm}
\begin{tabular}{c|c|c|c|c} 
 \hline
Scene & Object & Sequence & Train\# & Test\#  \\
\hline
1& 003$\_$cracker$\_$box & 0007 & 100 & 27 \\
2& 004$\_$sugar$\_$box & 0049 & 100 & 27 \\
3& 006$\_$mustard$\_$bottle & 0037 & 100 & 27 \\
4& 008$\_$pudding$\_$box & 0002 & 86 & 24 \\
5& 021$\_$bleach$\_$cleanser & 0036 & 100 & 27 \\
 \hline
\end{tabular}
\caption{Sequences and images used in YCB Video dataset.}
\label{tab:ycb}
\end{table}

\textbf{Tanks and Temples Dataset:}
For the Tanks and Temples dataset~\cite{knapitsch2017tanks}, we directly use the data published by~\cite{zhang2020nerf++} for fair comparisons. We use the Truck scene to validate our performance on object-centered unbounded scenes. We also manually label all the testing images for no ground truth masks are provided. Since the goal of our method is to decompose the object from backgrounds, we manually remove the meshes of the ground surface in the ground truth mesh file, and use this edited mesh when evaluating CD metric for all experiments.

\textbf{BlendedMVS Dataset:}
For the BlendedMVS dataset~\cite{yao2020blendedmvs}, we use the data  published by~\cite{wang2021neus} and \cite{wang2023autorecon} for fair comparisons. We use the Bear, Stone, and Gundam scene to validate our performance on more object-centered scenes. 
We select images with indices that are multiples of 10 from each sequence as the test set, and the rest are used as the training set.
We also manually label all the testing images for no ground truth masks provided. Since the goal of our method is to decompose the object from backgrounds, we manually remove the meshes of the ground surface in the ground truth mesh file, and use this edited mesh when evaluating CD metric for all experiments.

\begin{table*}[t]\small
\renewcommand{\arraystretch}{1.1}
\centering
\setlength{\tabcolsep}{3.3mm}
\begin{tabular}{l|c|cc|cc|ccc|c} 
 \hline
 Dataset & Scene ID & DINO-Coseg & SAM & SNeRF+D & DFF & NeuS & NeuS+D & NeuS+S & Ours \\
 \hline
 BlendedMVS & 1 & 0.595 & 0.614 & 0.643 & \slash & 0.772 & 0.643 & \textbf{0.875} & 0.784 \\
            & 2 & 0.830 & 0.570 & 0.720 & \slash & 0.523 & 0.720 & 0.869 & \textbf{0.879} \\
            & 3 & 0.843 & 0.437 & 0.879 & \slash & 0.475 & 0.879 & 0.683 & \textbf{0.964} \\
            & 4 & 0.819 & 0.686 & 0.738 & \slash & 0.512 & 0.849 & \textbf{0.867} & 0.862 \\
            & Mean & 0.772 & 0.577 & 0.745 & \slash  & 0.571 & 0.773 & 0.824 & \textbf{0.872} \\
 \hline
 Tanks Temples & 1 & 0.890 & 0.603 & 0.916 & \slash  & 0.558 & 0.838 & 0.603 & \textbf{0.946} \\
 \hline 
 DTU & 1 & 0.900 & 0.936 & 0.887 & \slash  & 0.792 & 0.861 & 0.949 & \textbf{0.971} \\
 \hline
 OnePose & 1 & 0.911 & 0.612 & 0.912 & 0.869 & 0.354 & 0.783 & 0.955 & \textbf{0.979} \\
        & 2 & 0.724 & 0.796 & 0.665 & 0.823 & 0.293 & 0.801 & 0.945 & \textbf{0.958} \\
        & 3 & 0.886 & 0.888 & 0.900 & 0.894 & 0.543 & 0.838 & 0.896 & \textbf{0.947} \\
        & Mean & 0.840 & 0.765 & 0.826 & 0.862  & 0.397 & 0.807 & 0.932 & \textbf{0.961} \\
 \hline 
 YCB Video & 1 & 0.605 & 0.743 & 0.662 & 0.707 & 0.353 & 0.591 & 0.952 & \textbf{0.955} \\
            & 2 & 0.702 & 0.843 & 0.746 & \textbf{0.938}  & 0.551 & 0.451 & 0.489 & 0.926 \\
            & 3 & 0.741 & 0.821 & 0.749 & 0.774 & 0.500 & 0.567 & 0.911 & \textbf{0.920} \\
            & 4 & 0.525 & 0.769 & 0.533 & 0.682 & 0.576 & 0.596 & \textbf{0.927} & 0.920 \\
            & 5 & 0.763 & 0.599 & 0.806 & 0.895 & 0.408 & 0.858 & \textbf{0.935} & 0.930 \\
            & 6 & 0.695 & 0.845 & 0.723 & 0.878 & 0.561 & 0.673 & 0.917 & \textbf{0.961} \\
            & Mean & 0.672 & 0.770 & 0.703 & 0.812 & 0.492 & 0.0.623 & 0.855 & \textbf{0.942} \\
 \hline 
\end{tabular}
\caption{Quantitative results of foreground segmentation performance with mIoU metric.}
\label{tab:seg}
\end{table*}

\begin{figure*}[htb]
\begin{center}
\includegraphics[width=0.98\linewidth]{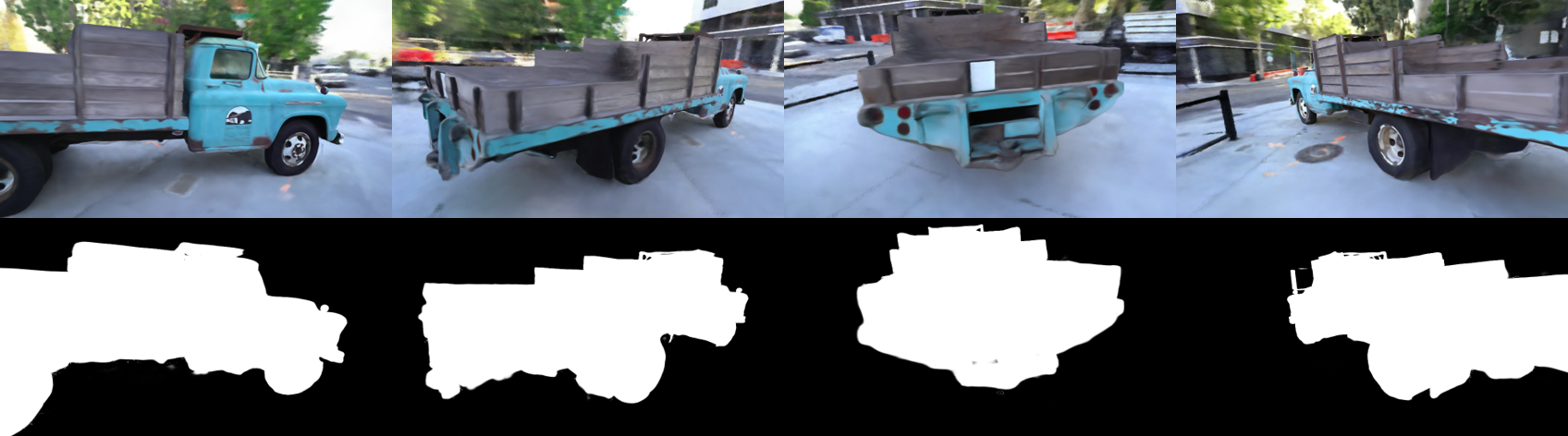}
\end{center}
   \caption{Additional qualitative results of novel view synthesis (top row) and foreground segmentation (bottom row) on the Tanks and Temples dataset.}
\label{fig:add-tt}
\end{figure*}

\textbf{YCB Video Dataset}
Most of the datasets commonly used in the reconstruction task are object-centered and non-cluttered, so we choose this dataset to validate our performance in cluttered, non-object-centered scenes.
The YCB Video dataset~\cite{xiang2017posecnn} is widely used in object pose estimation tasks. It contains 92 image sequences, in which objects sampled from 21 different categories are randomly placed or stacked on different tables. 
We use 5 objects from this dataset, and choose 1 sequence for each object. We downsample the original image sequences with a ratio of 20 to get training images, and randomly sample from the rest to get testing images. 
Tab.~\ref{tab:ycb} shows the specific scenes, objects, sequences, and number of images we use for training and testing.
We use the ground truth mask it provides for segmentation results evaluation.
Though the dataset provides poses of the objects with respect to cameras in each view, we observe pose errors that lead to multi-view inconsistency. 
Hence, we adopt COLMAP~\cite{schonberger2016structure} to refine the poses
and obtain the sparse pointcloud for sphere annotation following~\cite{wang2021neus}. After pose refinement, the coordinates we use differ from the coordinate the ground truth mesh use. To evaluate the reconstruction accuracy with the provided ground truth mesh, we adopt a trajectory registration
algorithm~\cite{umeyama1991least} to align our reconstructed coordinate with the ground truth coordinates. The aligned transformation is shared among all methods for fair comparisons.

\subsection{Additional Quantitative Results}

\textbf{Additional quantitative results on comparison to the state-of-the-art.}
In Tab.~\ref{tab:seg} we report the full results of the comparison to the state-of-the-art in foreground segmentation.

\subsection{Additional Qualitative Results}

\begin{figure*}[htb]
\begin{center}
\includegraphics[width=0.95\linewidth]{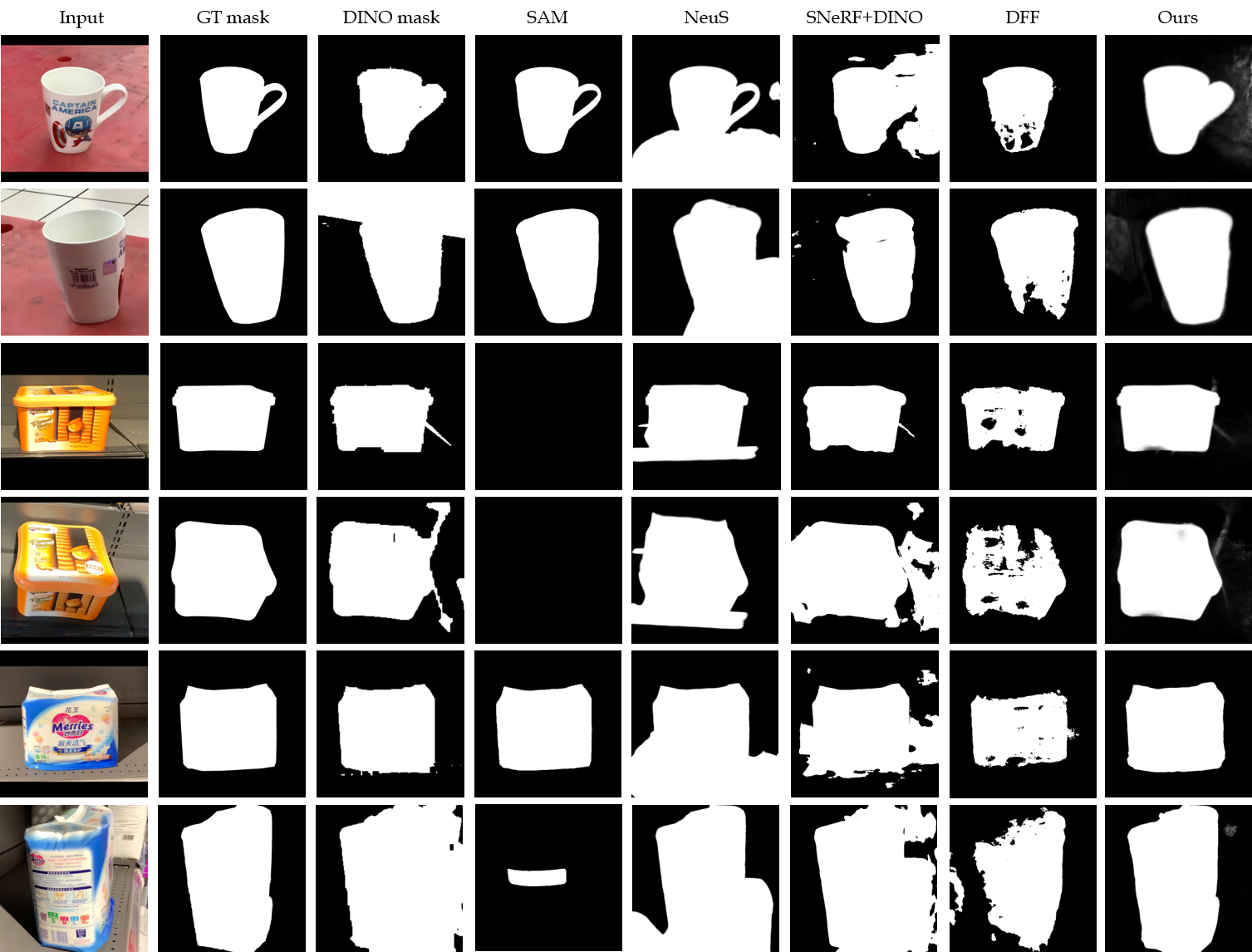}
\end{center}
   \caption{Additional qualitative segmentation results on the OnePose dataset.}
\label{fig:add-onepose-seg}
\end{figure*}

\begin{figure*}[htb]
\begin{center}
\includegraphics[width=0.98\linewidth]{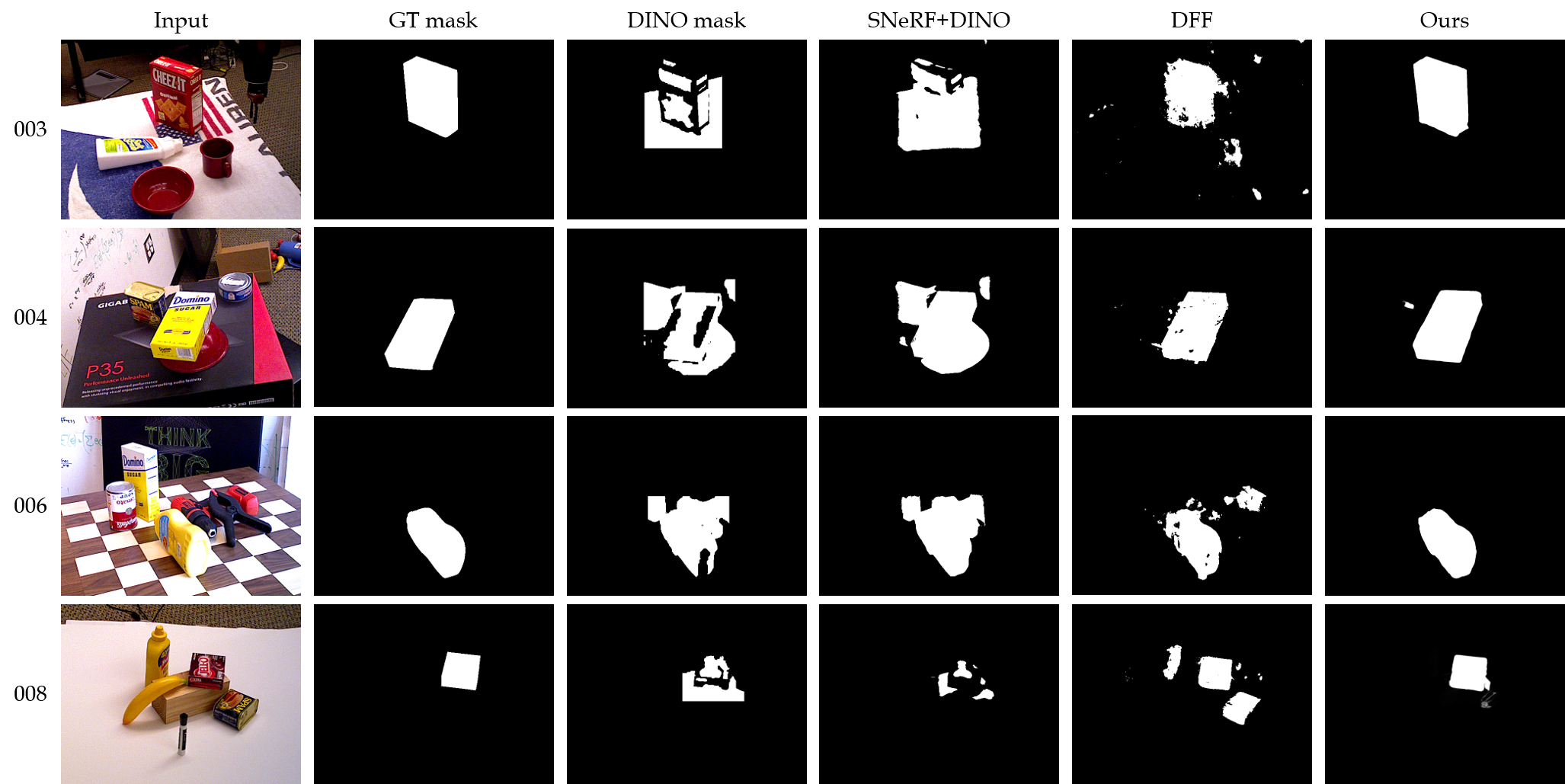}
\end{center}
   \caption{Additional qualitative segmentation results on the YCB Video dataset.}
\label{fig:add-ycb-seg}
\end{figure*}

\textbf{Additional qualitative results on foreground segmentation.}
We show additional qualitative results on foreground segmentation in Fig.~\ref{fig:add-tt} Fig.~\ref{fig:add-onepose-seg}, and Fig.~\ref{fig:add-ycb-seg}.

\begin{figure*}[htb]
\begin{center}
\includegraphics[width=0.9\linewidth]{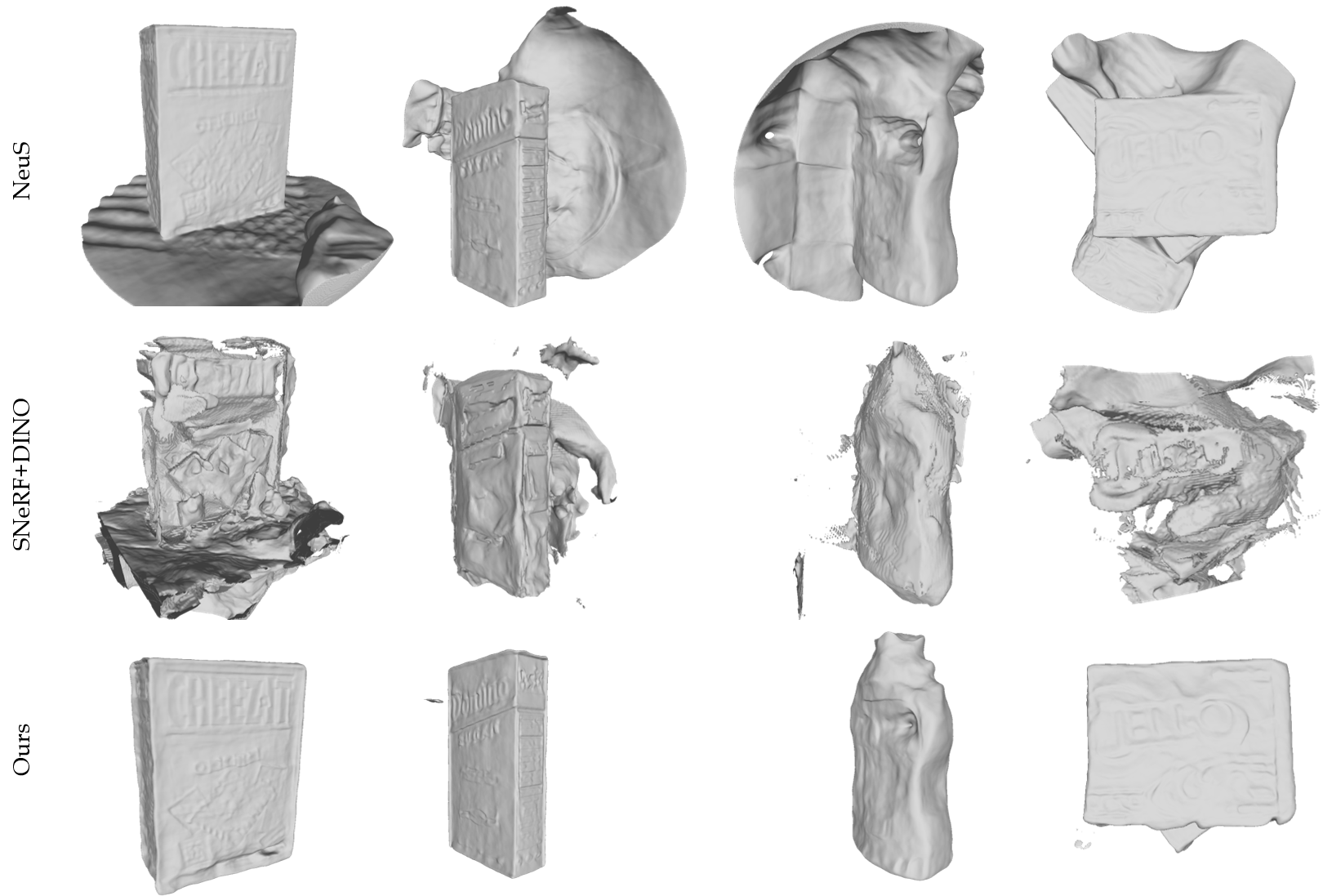}
\end{center}
   \caption{Additional qualitative reconstrution results on the YCB Video dataset.}
\label{fig:add-ycb-recon}
\end{figure*}

\begin{figure*}[htb]
\begin{center}
\includegraphics[width=\linewidth]{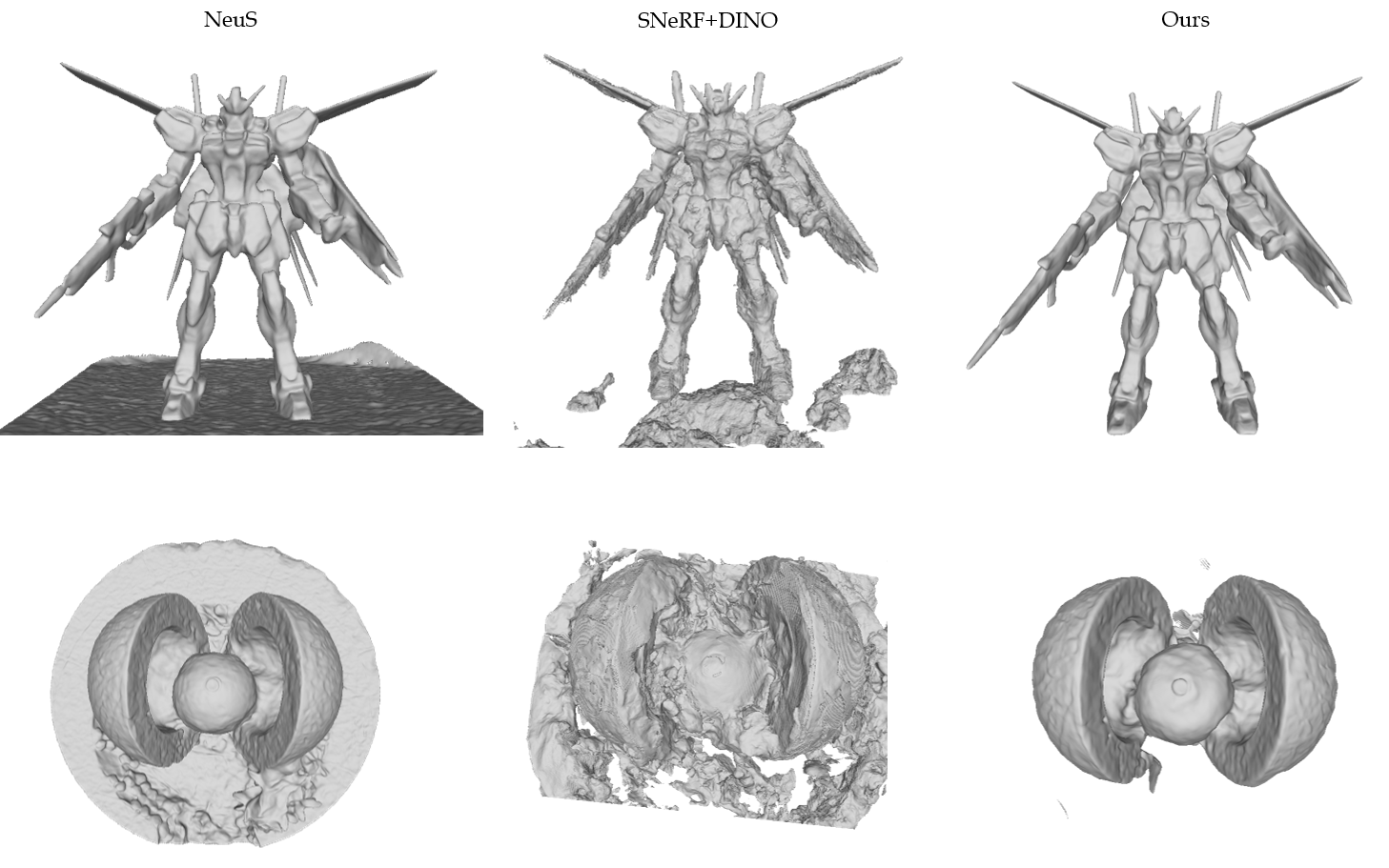}
\end{center}
   \caption{Additional qualitative reconstruction results on the BlendedMVS dataset.}
\label{fig:add-bmvs-recon}
\end{figure*}

\begin{figure*}[htb]
\begin{center}
\includegraphics[width=\linewidth]{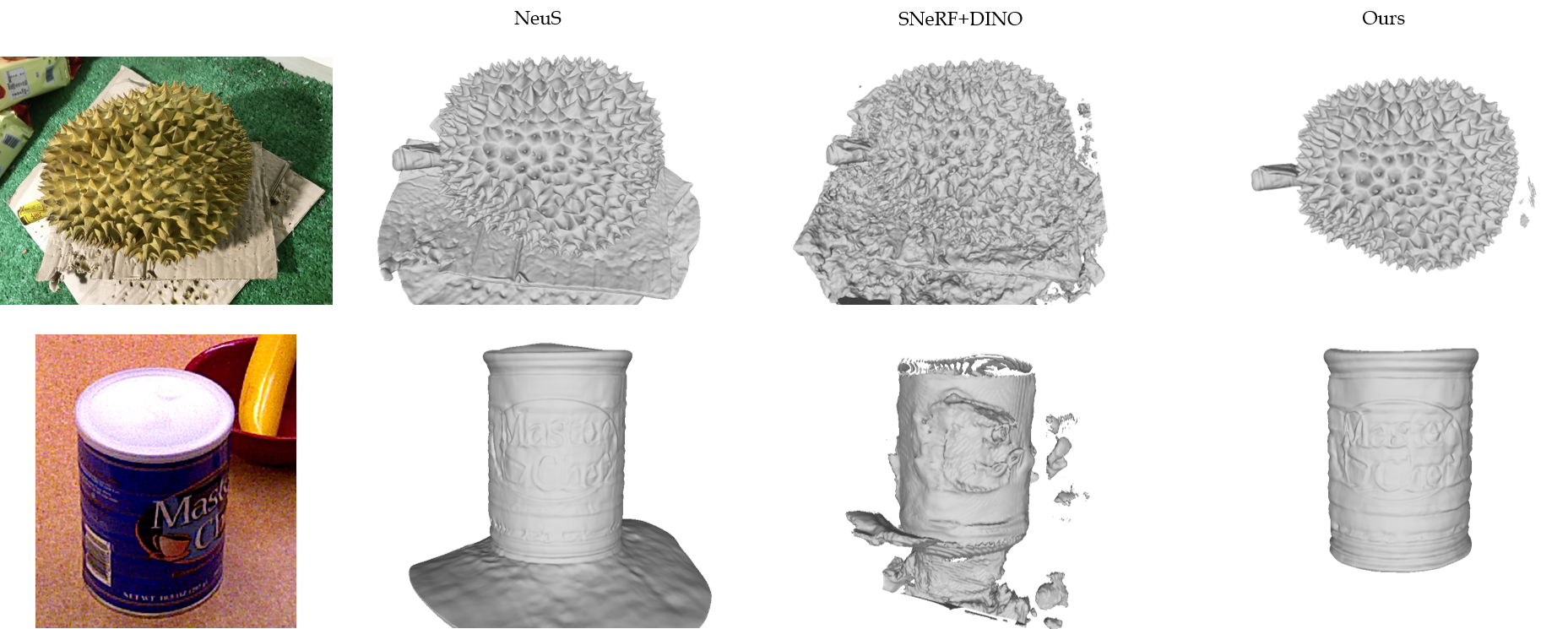}
\end{center}
   \caption{Additional qualitative reconstruction results.}
\label{fig:add-more-recon}
\end{figure*}

\textbf{Additional qualitative results on foreground reconstruction.}
We show additional qualitative results on foreground reconstruction in Fig.~\ref{fig:add-ycb-recon}, Fig.~\ref{fig:add-bmvs-recon}, and Fig.~\ref{fig:add-more-recon}.

\subsection{Limitations}\label{sec4}
Currently, our method still holds two limitations. First, when part of the background is encased by the target object and keeps consistent across views, we are unable to separate it correctly, e.g. the red table circled by the handle of the captain cup object in OnePose dataset. Second, the training process still takes hours, and we plan to accelerate it with advanced network architectures or training strategies in the future.

\end{appendix}

\end{document}